%% file: ms.tex
\documentclass{amia}
\usepackage{graphicx}
\usepackage[labelfont=bf]{caption}
\usepackage[superscript,nomove]{cite}
\usepackage{color}

\usepackage{subfiles}

\newif\ifsubfile
\subfiletrue

\usepackage{soul}
\usepackage[color=yellow, textsize=small]{todonotes}
\usepackage{wrapfig}
\usepackage{multirow}
\usepackage{array}
\usepackage{booktabs}
\usepackage[hidelinks]{hyperref}
\usepackage{url}
\usepackage{enumitem}
\makeatletter
\renewcommand{\@biblabel}[1]{\hfill #1.}
\makeatother
\usepackage{amsmath}
\usepackage{bm}
\usepackage{subcaption}
\usepackage{graphicx}
\usepackage[export]{adjustbox}

\let\oldbibliography\thebibliography
\renewcommand{\thebibliography}[1]{%
  \oldbibliography{#1}%
  \setlength{\itemsep}{-0.1em}%
}

\begin{document}
\subfilefalse

\title{Identifying ARDS using the Hierarchical Attention Network with Sentence Objectives Framework}

\author{Kevin Lybarger, PhD, Linzee Mabrey, MD, Matthew Thau, MD, \\ Pavan K. Bhatraju, MD, MSc, Mark Wurfel, MD, PhD, Meliha Yetisgen, PhD}

\institutes{
    University of Washington, Seattle, WA, USA\\
}

\maketitle

\noindent{\bf Abstract}

\subfile{sections/abstract}

\section*{Introduction}

\subfile{sections/introduction}

\section*{Related Work}

\subfile{sections/related_work}

\section*{Methods}

\subfile{sections/methods}

\section*{Results}
\subfile{sections/results}

\section*{Conclusions}

\subfile{sections/conclusions}

\section*{Acknowledgements}
This work was supported by NIH/NHGRI (1U01 HG-008657), NIH/NLM Biomedical and Health Informatics Training Program (5T15LM007442-19), and NIDDK K23DK116967 (PKB). We want to acknowledge Sudhakar Pipavath, MD, for his contributions to the annotation of the chest radiograph images used in this study. Research and results reported in this publication was partially facilitated by the generous contribution of computational resources from the University of Washington Department of Radiology.

\bibliography{mybib}

\end{document}

%% file: sections/abstract.tex
\textit{Acute respiratory distress syndrome (ARDS) is a life-threatening condition that is often undiagnosed or diagnosed late. ARDS is especially prominent in those infected with COVID-19. We explore the automatic identification of ARDS indicators and confounding factors in free-text chest radiograph reports. We present a new annotated corpus of chest radiograph reports and introduce the Hierarchical Attention Network with Sentence Objectives (HANSO) text classification framework. HANSO utilizes fine-grained annotations to improve document classification performance. HANSO can extract ARDS-related information with high performance by leveraging relation annotations, even if the annotated spans are noisy. Using annotated chest radiograph images as a gold standard, HANSO identifies bilateral infiltrates, an indicator of ARDS, in chest radiograph reports with performance (0.87 F1) comparable to human annotations (0.84 F1). This algorithm could facilitate more efficient and expeditious identification of ARDS by clinicians and researchers and contribute to the development of new therapies to improve patient care.}

%% file: sections/introduction.tex
Coronavirus disease 2019 (COVID-19) is caused by infection with the severe acute respiratory syndrome coronavirus-2 (SARS-CoV-2) and is associated with high mortality.\cite{bhatraju2020covid} A high-risk complication of COVID-19 infection is the development of the acute respiratory distress syndrome (ARDS), which is characterized by severe inflammatory lung injury. Other common hospital diagnoses, such as sepsis, pneumonia, and trauma, are also associated with the development of ARDS. Interventions to prevent injury from invasive mechanical ventilation and differences in clinical management have improved clinical outcomes in patients with ARDS;\cite{acute2000ventilation, national2006comparison, guerin2013prone} however, ARDS is commonly under recognized by clinicians. In an epidemiologic study involving 500 intensive care units across 50 countries, over 40\% of all ARDS cases were not recognized by clinicians, and the diagnosis of over 30\% of ARDS cases was delayed.\cite{bellani2016epidemiology} In another study, investigators demonstrated that delays in initiating evidence-based treatments was associated with increased hospital mortality in patients with ARDS.\cite{needham2015timing} 

The identification of ARDS requires the assessment of lung injury patterns in chest imaging. A primary contributor to undiagnosed ARDS is the challenge of incorporating radiologist-derived chest imaging findings into diagnostic algorithms for ARDS. Per the ``Berlin Definition,'' ARDS diagnosis requires:\cite{force2012acute}
\begin{itemize}[nosep, leftmargin=0.13in]
    \item \textit{timing}: condition occurs within one week of a known clinical insult or new/worsening respiratory symptoms
    \item \textit{chest imaging}: bilateral opacities that are not fully explained by effusions, lobar or lung collapse, nodules, or masses 
    \item \textit{non-cardiogenic edema}: alveolar infiltrates are not fully explained by cardiac failure or hydrostatic edema
    \item \textit{oxygenation}: oxygenation measurements meet defined thresholds (mild, moderate, and severe)
\end{itemize}
The \textit{oxygenation} component requires decreased oxygenation and is generally documented in structured data in the electronic health record (EHR). The \textit{non-cardiogenic edema} component requires an absence of hydrostatic edema, and the associated risk factors may be captured in structured admit diagnosis codes or the clinical narrative. The \textit{timing} component requires a proximal risk factor for respiratory failure, for example the presence of COVID-19, and may be documented through lab results or diagnosis codes. The information needed to assess the \textit{chest imaging} requirements is typically represented in chest radiographs (x-rays) and computed tomography images, as well as the associated free-text reports describing radiologists' findings and interpretation. Data-driven computer vision approaches for directly analyzing the chest radiographs images are still in development and are computationally expensive. This work explores the automatic identification of the \textit{chest imaging} requirements for ARDS in free-text chest radiograph reports. 

We use natural language processing (NLP) information extraction techniques to identify descriptions of opacities (increased radiodensity), classify the opacities as parenchymal (indicative of alveolar edema/infiltrates) or extraparenchymal (outside the lungs or not indicative of alveolar edema/infiltrates), resolve sidedness (unilateral or bilateral), capture size information (small, moderate, or large), and indicate negation (``not present''). We developed detailed annotation guidelines that include summary document-level annotations and detailed relation annotations that characterize opacities. Using this novel annotation scheme, we created a new annotated corpus of 420 chest radiograph reports, referred to as the Pulmonologist Annotated Corpus (PAC). This work presents the Hierarchical Attention Network with Sentence Objectives (HANSO) framework, which is an end-to-end neural model that utilizes both the document-level and relation annotations. We introduce an approach for leveraging entity and relation annotations with noisy spans to improve document classification performance within the HANSO framework. We compare the performance of HANSO against two gold standards: manually annotated chest radiograph reports and manually annotated chest radiograph images. HANSO achieves very high performance in identifying the presence of bilateral infiltrates, a key indicator of ARDS, relative to both the annotated reports (0.87 F1) and annotated images (0.87 F1). HANSO also identifies factors that are less consistent for ARDS, specifically extraparenchymal opacities, with high performance (0.80 F1).

\ifsubfile
\bibliography{mybib}
\fi

%% file: sections/related_work.tex

Many works explore NLP information extraction techniques with radiology reports.\cite{SORIN2020639} Within this body of radiology research, several works explore the identification of pulmonary conditions in chest radiograph reports. Most prior pulmonary information extraction work implements discrete document classification models where labels are assigned at the document-level, without utilizing word-level annotations or predictions. Bejan, et al. identify pneumonia in chest radiograph reports using Support Vector Machines (SVM) with word n-grams, medical concepts, and other features.\cite{bejan2012pneumonia} Yetisgen, et al. automatically identify acute lung injury in chest radiograph reports using Maximum Entropy (MaxEnt) models that utilize word n-grams and assertion predictions (present vs. absent).\cite{yetisgen2013identification} Afshar et al. and Mayampurath et al. predict ARDS in chest radiograph reports  using word n-grams and medical concept features using discrete modeling approaches, including decision trees, k-nearest neighbors, naive bayes, logistic regression, and SVM.\cite{afshar2018computable, mayampurath2020external} Mayampurath et al. achieves the best performance in predicting ARDS using unigram term frequency–inverse document frequency (TF-IDF) features with SVM,\cite{mayampurath2020external} which we implement here as a baseline. 

Some recent NLP work with chest radiograph reports utilizes continuous, neural modeling approaches. Datta et al. annotate approximately 2,000 chest radiograph reports using a detailed relation-based annotation scheme that characterizes radiology phenomena across multiple dimensions. Datta implements neural entity and relation extraction models, including a baseline model consisting of stacked bidirectional long short-term memory (bi-LSTM) and conditional random field layers, as well as transformer-based approaches using BERT and XLNet.\cite{datta2020understanding} Apostolova et al. investigate ARDS using both clinical text and structured EHR data from the MIMIC-III database, exploring ARDS likelihood, mortality, and risk factors using learned vector patient representations.\cite{apostolova2019towards} Apostolova's patient vectors incorporate information from clinical notes, diagnosis codes, and other structured data using Convolutional Neural Networks and Gradient Boosting Machine.

This work is differentiated from prior ARDS-related information extraction work in multiple ways. This work presents a new detailed annotation scheme that identifies indicators and confounding factors for ARDS, including document-level summary labels and detailed relation annotations describing the support or evidence for the document-level labels. It introduces a new end-to-end, neural multitask model that predicts the document-level labels and utilizes the detailed relation annotations to augment learning. Additionally, this work presents an approach for leveraging noisy entity and relation annotations.

\ifsubfile
\bibliography{mybib}
\fi

%% file: sections/methods.tex
\subsection*{Data} 
This work utilized two existing clinical data sets from the University of Washington Harborview and Montlake campuses. The first data set, \textit{Data set A}, includes 831 chest radiograph reports for 173 patients from February-September 2020. \textit{Data set A} includes patients that were being evaluated under suspicion for COVID-19 and admitted to a medical or trauma intensive care unit (ICU). Inclusion criteria were: (1) ICU admission; (2) suspicion for COVID-19; (3) invasive mechanical ventilation; (4) presence of at least one partial pressure of arterial blood oxygen-to-fraction of inspired oxygen ratio (Pa02/FI02) less than 300 mmHg. For \textit{Data set A}, an expert chest radiologist annotated 154 radiograph images using the Berlin criteria to identify patients with diffuse bilateral pulmonary opacities. The second data set, \textit{Data set B}, includes 1,279 radiograph reports for 788 patients from March-November, 2020. \textit{Data set B} includes all patients hospitalized with COVID-19, resulting in a broader patient population than \textit{Data set A} with varying degrees of severity of illness from COVID-19 infection.

\begin{wrapfigure}{R}{3.0in}
  \begin{center}
    \includegraphics[width=2.9in, frame]{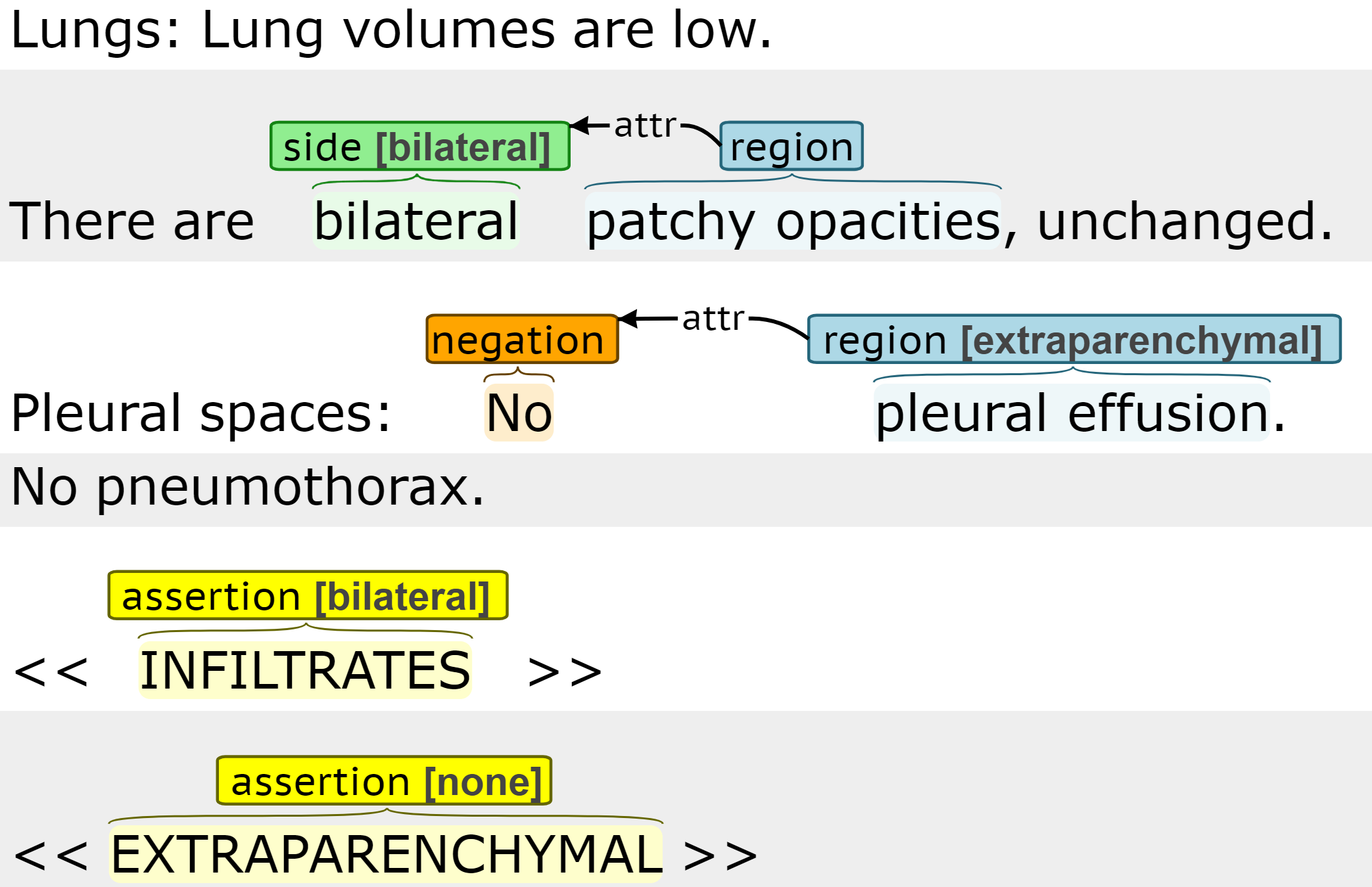}
  \end{center}
  \caption{BRAT annotation example}
  \label{brat_example}
\end{wrapfigure}

\textbf{Annotation Scheme:} We developed a detailed annotation scheme that facilitates the identification of lung infiltrates and extraparenchymal opacities. Table \ref{annotated_phenomena} summarizes the annotated phenomena, and Figure \ref{brat_example} presents annotation examples from the BRAT annotation tool.\cite{stenetorp2012brat} Each report was annotated with two categories of labels: document labels and relational labels.  The document labels summarize the annotators' overall assessment of each chest radiograph report with two multiclass labels: \textit{infiltrates} – consistent with ARDS and \textit{extraparenchymal} – less consistent with ARDS. The document classes are: \textit{none} – insufficient information for assessment or absence explicitly stated; \textit{present} – condition present but sidedness unknown; \textit{unilateral} – explicitly one lung; and \textit{bilateral} – explicitly both lungs. The string, ``$\langle\langle$ INFILTRATES $\rangle \rangle$ $\langle\langle$ EXTRAPARENCHYMAL $\rangle \rangle$,'' was appended to each report to facilitate the assignment of these document labels.

The relation annotations are evidence for the document labels and include annotated spans and links between spans. Although the annotated spans are not necessarily noun phrases, we refer to the spans as ``entities'' here. The entity types include \textit{region}, \textit{side}, \textit{size}, and \textit{negation}. The annotation of \textit{region}, \textit{side}, and \textit{size} includes an identified span and the assignment of a subtype label that normalizes the span contents, mapping the phrase to a clinically significant label. For example, all \textit{region} entities include a subtype label of \textit{parenchymal} or \textit{extraparenchymal}. The \textit{negation} entity only includes an annotated span without a subtype label, although the type label conveys the span meaning (i.e. ``absent''). The relation annotations indicate whether a \textit{side}, \textit{size}, or \textit{negation} entity are an attribute of a \textit{region} entity. All attribute (\textit{attr}) relation annotations are unidirectional, where the first entity in all relations has type \textit{region}).

The annotation scheme provides the information necessary to categorize each chest radiograph report with respect to the radiologic criteria for ARDS; namely the presence of bilateral opacities that are not fully explained by effusions, lobar or lung collapse, nodules, or masses. The presence of opacities was qualified as \textit{infiltrates} (indicative of alveolar process) and/or \textit{extraparenchymal} (indicative of effusions, collapse, nodules/masses, and atelectasis) with additional annotation of any report text documenting laterality and size. This approach mirrors the clinical heuristic used by expert radiologists and pulmonary/critical care clinicians when assessing the likelihood that a chest radiograph indicates the presence of ARDS. In our annotation scheme the document labels, \textit{infiltrates} and \textit{extraparenchymal}, are the best indicators of ARDS, and the relation annotations are included to support the document labels.

\begin{table}[ht]
    \small
    \centering

\input{tables/annotated_phenomena.tex}
    \caption{Annotation guideline summary}
    \label{annotated_phenomena}
\end{table}

\textbf{Annotation Scoring and Evaluation:} The goal of this work is to extract salient information from chest radiograph reports and convert it to a structured representation that will complement other types of structured clinical data (e.g. PaO3/FiO2 ratio) to predict ARDS. The assessment of annotator agreement and extraction performance focuses on the information in the annotation schema that is most relevant to the large-scale, automated assessment of ARDS. For each entity, the subtype label captures the important span information, such that the associated text span is less informative. 
\begin{wrapfigure}{r}{3.6in}
  \begin{center}
    \includegraphics[width=3.5in, frame]{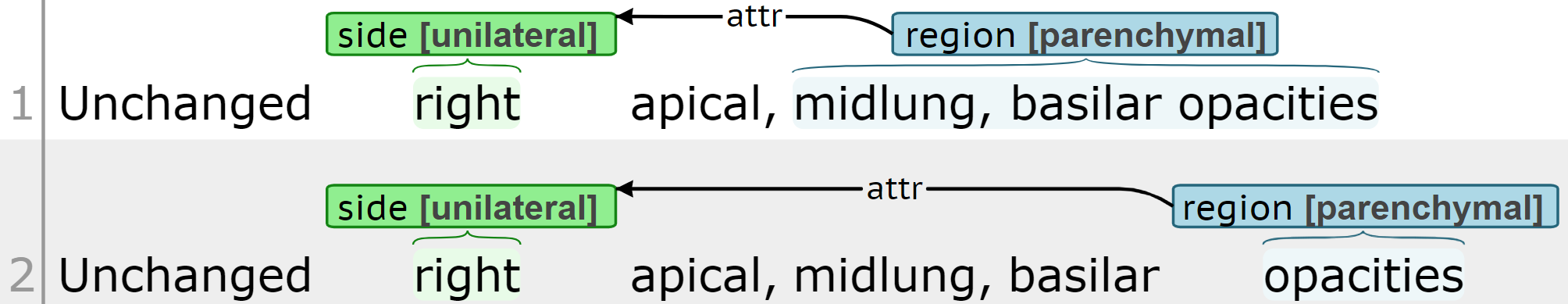}
  \end{center}
  \caption{Annotation comparison example}
  \label{slot_filling}
\end{wrapfigure}
Figure \ref{slot_filling} presents the same sentence annotated by two annotators. Both annotators label a \textit{region} entity with subtype \textit{parenchymal} that is connected to a \textit{side} entity with subtype \textit{unilateral}. Although the annotators label different spans for the \textit{region} entities (``midlung, basilar opacities'' vs. ``opacities''), both annotations identify unilateral parenchymal opacities (i.e. opacities in one lung). For the purposes of predicting ARDS, these annotations are equivalent, even though there are span differences. For entities with equivalent type and subtype labels, the spans are evaluated under two criteria: \textit{any overlap} and \textit{partial match}. Under the \textit{any overlap} criterion, spans are considered equivalent if there is at least one overlapping token, and the performance is assessed based on span counts. For the \textit{region} annotation in Figure \ref{slot_filling}, the entity spans ``midlung, basilar opacities'' and ``opacities'' overlap, so there is one matching span. Under the \textit{partial match} criterion, spans are compared at the token level to allow partial matches, and performance is assessed based on the number of matching tokens. For the \textit{region} annotation in Figure \ref{slot_filling}, the entity spans have one matching token (``opacities'') and three mismatched tokens (``midlung , basilar''). There is only one relation type (\textit{attribute} or \textit{attr}), and two relations are equivalent if the entities paired by the \textit{attribute} relation are equivalent under the \textit{any overlap} criterion. Performance is evaluated using precision (P), recall (R), and F1-score (F1). 

\begin{wrapfigure}{R}{2.5in}
  \begin{center}
    \includegraphics[width=2.3in]{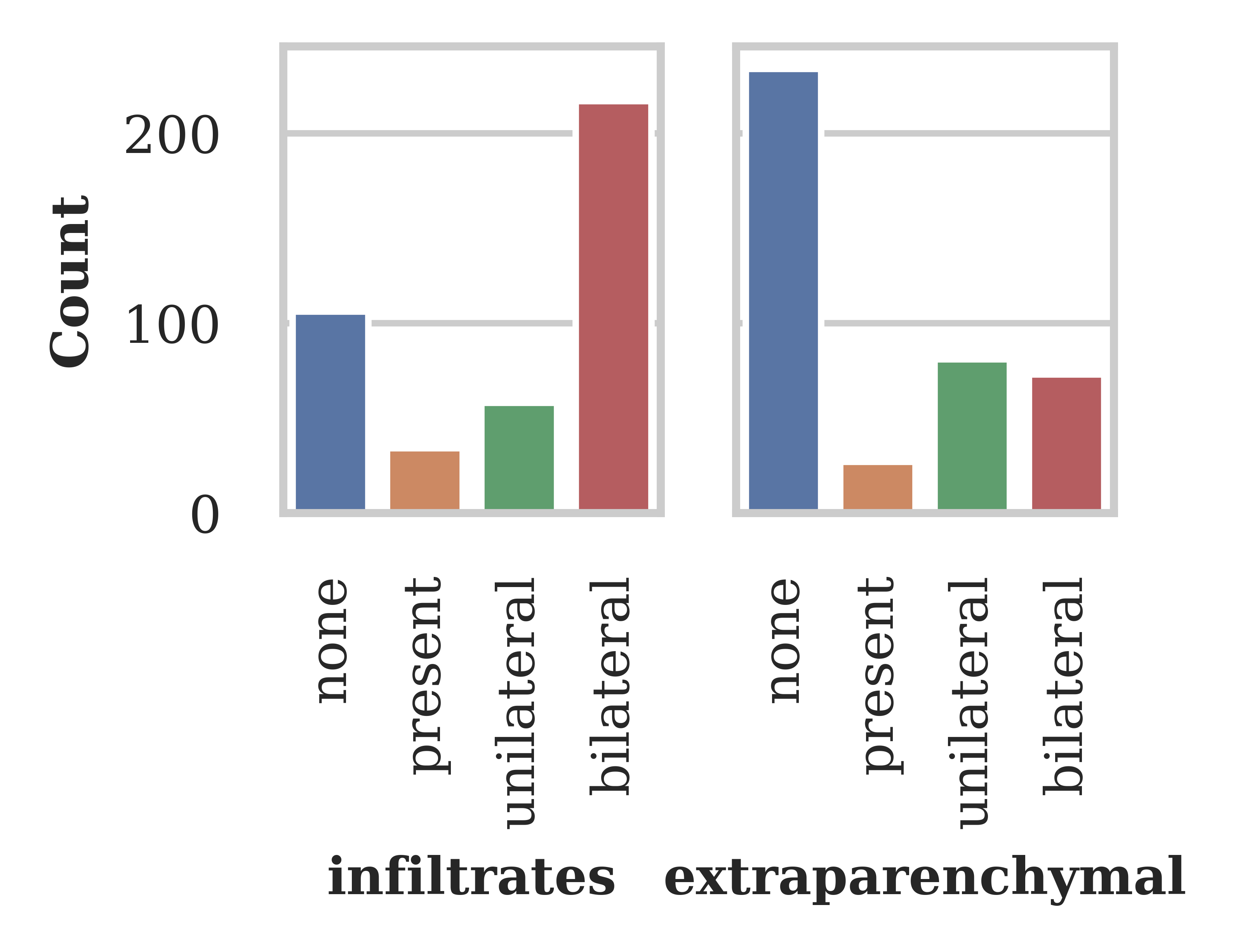}
  \end{center}
  \caption{Document label histogram}
  \label{document_label_histogram}
\end{wrapfigure}

\textbf{Annotation Statistics:} The Pulmonologist Annotated Corpus (PAC) includes 420 chest radiograph reports, 120 from \textit{Data set A} and 300 from \textit{Data set B}. PAC was annotated by two pulmonary and critical care fellows, whom each annotated half the corpus. PAC has an 80\%/20\% train/test split. Figure \ref{document_label_histogram} presents the histogram of the PAC document labels. The indication or reason for a chest radiograph in this patient population is a respiratory complaint, so positive labels ($label \in \{present, unilateral, bilateral\}$) are frequent: 75\% of reports have a positive \textit{infiltrates} label, and 43\% have a positive \textit{extraparenchymal} label. The corpus includes an average of 3.0 \textit{region}, 1.3 \textit{side}, 0.8 \textit{negation}, and 0.2 \textit{size} entities and 2.7 relations per report.

\textbf{Annotator Agreement:} As part of the annotation guideline development and annotator training, we doubly annotated 20 reports, assessed inter-annotator agreement, updated the annotation guidelines, and provided additional annotator training. At the conclusion of the project, we doubly annotated 10 additional reports, to assess the agreement for the annotated corpus The agreement for these 10 reports is presented in Table \ref{annotator_agreement}. The agreement for both document labels is high (0.90 F1). The entity agreement under the \textit{any overlap} criteria is also high (0.85-0.96 F1). The entity agreement using the \textit{partial match} criteria remains high for \textit{negation} and \textit{side} entities; however it is relatively low for \textit{region}. These results suggest the annotators are generally labeling the same phenomenon; however, they differ in the selected spans, similar to the example in Figure \ref{slot_filling}. Relation agreement is very high for \textit{region}-\textit{negation} entity pairs (0.96 F1) and lower for \textit{region}-\textit{side} pairs (0.76 F1). While the \textit{region} span annotations are noisy, the entity and relation annotations, still contain useful information for assessing ARDS. 

\begin{table}[ht]
    \small
    \centering
    \begin{subtable}[t]{1.9in}    
        \centering

\input{tables/agreement_doc.tex}
        \caption{Document labels}
    \end{subtable}
    \begin{subtable}[t]{2.6in}    
        \centering

\input{tables/agreement_entity.tex}
        \caption{Entities}
    \end{subtable}
    \begin{subtable}[t]{1.9in}    
        \centering

\input{tables/agreement_relation.tex}
        \caption{Relations}
    \end{subtable}
    \caption{Annotator agreement. $^*$no size spans were annotated.}
    \label{annotator_agreement}    
\end{table}

\subsection*{Information Extraction}
\begin{wrapfigure}{R}{2.9in}
  \begin{center}
    \includegraphics[width=2.8in]{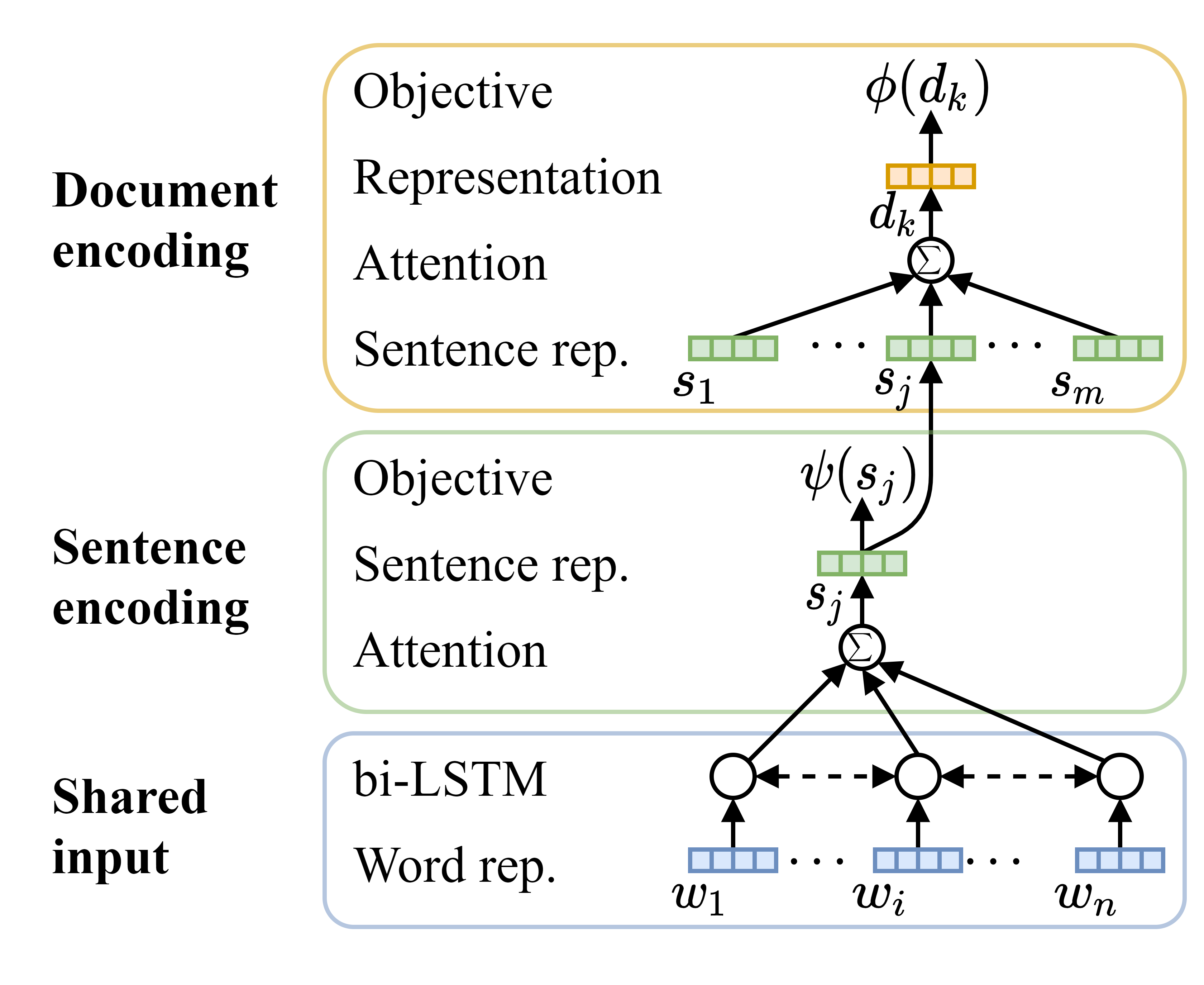}
  \end{center}
  \caption{HANSO framework}
  \label{hanso}
\end{wrapfigure}

The likelihood of a patient satisfying the \textit{chest imaging} requirements for the Berlin definition of ARDS can be estimated from the document labels, \textit{infiltrates} and \textit{extraparenchymal}. We introduce the Hierarchical Attention Network with Sentence Objectives (HANSO)  framework in Figure \ref{hanso}, to predict these document labels and incorporate the relation annotation information. HANSO is an neural end-to-end, multi-task model that includes sentence encoding and document encoding layers. It builds on Yang's hierarchical attention network (HAN).\cite{yang2016hierarchical} HAN aggregates word-level information to the sentence-level and then aggregates sentence-level information to the document-level. Sentences are encoded using a multi-layer network consisting of a recurrent neural network (RNN) and self-attention, and documents are encoded using separate RNN and self-attention layers operating on the encoded sentences. We build on HAN, incorporating sentence-level prediction tasks to augment the learning of the sentence representations. The sentence targets are derived from the relation annotations. The sentence and document encoding layers in Figure \ref{hanso} are implemented for each of the document labels, \textit{infiltrates} and \textit{extraparenchymal}, with a shared input recurrent layer. HANSO omits the additional RNN included in the document encoding layer of HAN.

In our initial experimentation, we implemented a span-based relation extraction model for the entity and relation labels, similar to our previous work,\cite{COVID_jbi_2020} and tried using the extracted relation information to improve the prediction of the document labels. However, the entity and relation extraction performance was insufficient to improve the prediction of the document labels, which are the most important labels in our annotation schema. Contributing factors to the low entity and relation extraction performance include the small data set size and the variability in the annotated spans, especially for \textit{region} entities. The subtype labels in our annotation scheme (e.g. \textit{unilateral} and \textit{bilateral} for the \textit{side} entities) normalize the span information, so the noisiness in the annotated spans does not negatively impact the informativeness of the annotations. However, this noise in the span annotation does negatively impact model learning and span prediction.  

\begin{figure}[ht]
  \begin{center}
    \includegraphics[width=5.6in]{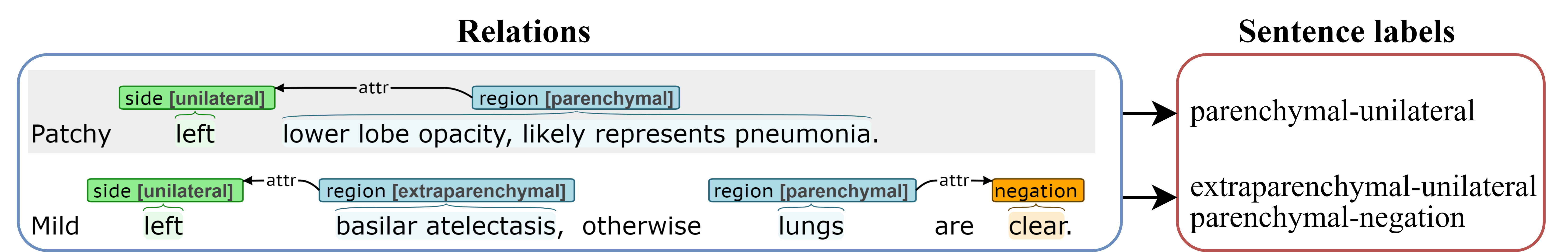}
  \end{center}
  \caption{Relation-to-sentence label mapping example}
  \label{sent_labels}
\end{figure}

To utilize the detailed relation information, the relations are converted to a one-hot encoding for each sentence, capturing the salient relation and entity information without explicitly identifying entity spans. Figure \ref{sent_labels} presents examples of this relations-to-sentence label mapping process. Relations consisting of  \textit{region}-\textit{side} pairs are represented as the subtype label pairs: \{\textit{parenchymal}-\textit{unilateral}, \textit{parenchymal}-\textit{bilateral}, \textit{extraparenchymal}-\textit{unilateral}, \textit{extraparenchymal}-\textit{bilateral}\}. Relations consisting of \textit{region}-\textit{negation} pairs are represented as the \textit{region} subtype label and \textit{negation} type: \{\textit{parenchymal}-\textit{negation}, \textit{extraparenchymal}-\textit{negation}\}. While this approach does not explicitly capture span information, this sentence-level encoding of the relations creates a summary of the most important annotated phenomena within each sentence. The \textit{size} entities are infrequent within the corpus and are omitted from experimentation. 

In the description of the HANSO framework below, the subscripts $i$, $j$, and $k$ indicate the $i^{th}$ BERT word piece position, $j^{th}$ sentence, and $k^{th}$ document. We only include the $i$, $j$, and $k$ subscripts below that are needed to resolve ambiguity.

\textbf{Input encoding:} Input documents are split into sentences and tokenized using spaCy.\cite{spacy} The default \textit{en\_core\_web\_sm} spaCy configuration is used, except that line breaks always indicate a sentence boundary. Sentences are mapped to contextualized word embeddings using \textit{Bio+Clinical BERT}.\cite{alsentzer-etal-2019-publicly} The contextualized BERT word piece embeddings feed into a bi-LSTM without fine tuning BERT (no backpropagation to BERT). We tried several different architectures involving fine tuning BERT but these architectures did not out perform HANSO, which is likely due to the very small training set. The forward and backward states of the bi-LSTM are concatenated to form $\bm{h}_i$ with size $1 \times l_h$, where $i$ is the word piece position.

\textbf{Sentence encoding:}  Each sentence is represented as the attention-weighted sum of the word piece vectors. The bi-LSTM hidden state for each word piece is nonlinearly projected from size $l_h$ to $l_p$ as
\begin{equation}
\bm{u}_{i} = tanh(\bm{W}_u \bm{h}_{i} + \bm{b}_u),
\label{attention_projection}
\end{equation}
where $\bm{W}_u$ is weight matrix and $\bm{b}_u$ is a bias vector. 
Word-level attention weights, $\bm{\alpha}_u$, are calculated using dot product attention as
\begin{equation}
\bm{\alpha}_{u, i} =\frac{\mbox{exp}(\bm{u}_i^T \bm{z}_u)}{\sum\limits_{i} \mbox{exp}(\bm{u}_i^T \bm{z}_u)},
\label{attention_alphas}
\end{equation}
where $\bm{u}_i$ is the projected word piece input and $\bm{z}_u$ is a learned vector of size $l_p$. The representation of sentence $j$ is calculated as
\begin{equation}
\bm{s}_j = \sum\limits_{i} \bm{\alpha}_{u, i}\bm{h}_i,
\label{attention_sum}
\end{equation}
where $\bm{s}_j$ has size $l_h$. A set of binary sentence-level prediction tasks, $R$, are incorporated based on the one-hot encoding of the relations described above. For each sentence-level task, $r \in R$, the sentence vector is nonlinearly projected from size $l_h$ to $l_p$ as 
\begin{equation}
    \bm{v}_{r,j} = tanh(\bm{W}_{v,r} \bm{s}_{j} + \bm{b}_{v,r}).
    \label{sent_objective_nonlin}
\end{equation}

Label scores for task $r$ are calculated using a linear projection from size $l_p$ to $2$ as 
\begin{equation}
    \bm{\psi}_{r,j} = \bm{W}_{\psi,r} \bm{v}_{r,j} + \bm{b}_{\psi,r},
    \label{sent_objective_lin}
\end{equation}
where $\bm{\psi}_{r,j}$ are the label scores for task $r$ and sentence $j$.

\textbf{Document encoding:} The same attention framework defined in Equations \ref{attention_projection} and \ref{attention_alphas} is used to calculate the sentence-level attention weights, $\bm{\alpha}_{s}$. Separate attention weights and bias vectors are learned for the sentence and document encoders. The sentence representation is nonlinearly projected from size $l_h$ to $l_p$ as 
\begin{equation}
\bm{x}_j = tanh(\bm{W}_x \bm{s}_j + \bm{b}_x).
\label{sent_nonlin}
\end{equation}
Each document is represented as the attention weighted sum of the projected sentence vectors as
\begin{equation}
\bm{d}_k = \sum\limits_{j} \bm{\alpha}_{s, j}\bm{x}_j,
\end{equation}
where $\bm{\alpha}_{s,j}$ is the attention weight for sentence $j$ and $\bm{d}_k$ has size $l_p$. The classes for the \textit{infiltrates} and \textit{extraparenchymal} labels are $\{none, present, unilateral, bilateral\}$. Document label predictions are generated by linearly projecting the document vector from size $l_p$ to $l_d$ as
\begin{equation}
\bm{\phi}_{k} = \bm{W}_d \bm{d}_{k} + \bm{b}_d
\end{equation}
where $\bm{\phi}_{k}$ are the label scores for document $k$ and $l_d$ is the document label set size.

Separate sentence and document encoders are implemented for the \textit{infiltrates} and \textit{extraparenchymal} document labels, utilizing a shared bi-LSTM input layer. For the prediction of the \textit{infiltrates} document label, the binary sentence prediction targets include: \textit{parenchymal}-\textit{unilateral}, \textit{parenchymal}-\textit{bilateral}, and \textit{parenchymal}-\textit{negation}. For the prediction of the \textit{extraparenchymal} document label, the binary sentence prediction targets include: \textit{extraparenchymal}-\textit{unilateral}, \textit{extraparenchymal}-\textit{bilateral}, and \textit{extraparenchymal}-\textit{negation}. To assess the contributions of the sentence-level objectives to document prediction performance, we implement HANSO without the sentence-level learning objective in Equation \ref{sent_objective_lin} (``HANSO lite'') and the full HANSO model (``HANSO full'').

\textbf{Baseline:} As a baseline for prediction performance, we implement a SVM model with unigram TF-IDF features, similar to Mayampurath et al.'s recent work predicting ARDS in chest radiograph reports.\cite{mayampurath2020external}

\textbf{Experimental Setup:} Model hyperparameters are tuned through 3-fold cross validation on the PAC training set. The best performing models are applied to the withheld PAC test set. HANSO is implemented in PyTorch, and the SVM is implemented with scikit-learn.\cite{NEURIPS2019_9015, scikit-learn} The HANSO configuration includes layer normalization at the bi-LSTM output and dropout ($do$) after the bi-LSTM and nonlinear projections in Equations \ref{sent_objective_nonlin} and \ref{sent_nonlin}. Additional hyperparameters include the number of epochs ($ne$), batch size ($bs$ as document count), learning rate (0.002), maximum document length (35 sentences), maximum sentence length (30 tokens), hidden size ($l_h=100$), and projection size ($l_p=100$). The Adam optimizer is used, loss is summed across all targets, and the gradient norm is clipped ($g_{max}=1.0$). For \textit{HANSO lite}, $do=0.1$, $ne=400$, and $bs=40$. For \textit{HANSO full}, $do=0.2$, $ne=150$, and $bs=10$. The SVM configuration includes the kernel (``rbf''), regularization ($C=10.0$), and convergence tolerance (0.001).


\ifsubfile
\bibliography{mybib}
\fi

%% file: tables/annotated_phenomena.tex
\begin{tabular}{p{0.55in} m{0.85in} m{2.2in} m{2.2in} }
\toprule
\textbf{Category}                              & \textbf{Type}             & \textbf{Subtypes}                                & \textbf{Span examples}                                 \\ \midrule
\multirow{2}{*}{Document}  & infiltrates          & none, present, unilateral, bilateral & --                                            \\ \cmidrule{2-4} 
                                        & extraparenchymal  & none, present, unilateral, bilateral & --                                            \\ \midrule
\multirow{4}{*}{Relational}               & region            & parenchymal, extraparenchymal                 & ``pulmonary infiltrates'' or ``lung disease'' \\ \cmidrule{2-4} 
                                        & side              & unilateral, bilateral                         & ``right'' or ``both sides''                   \\ \cmidrule{2-4} 
                                        & size              & small, moderate, large                        & ``trace'' or ``small''                        \\ \cmidrule{2-4} 
                                        & negation          & --                                            & ``not present'' or ``no''                     \\ \bottomrule     
\end{tabular}

%% file: tables/agreement_doc.tex
\newcolumntype{M}[1]{>{\centering\arraybackslash}m{#1}}

\begin{tabular}{m{1.0in} M{0.35in}}
\toprule
\textbf{Document label}         & \textbf{F1}  \\ \midrule
extraparenchymal                & 0.90         \\
infiltrates                     & 0.90         \\ \bottomrule
\end{tabular}

%% file: tables/agreement_entity.tex
\newcolumntype{M}[1]{>{\centering\arraybackslash}m{#1}}

\begin{tabular}{m{0.40in} M{0.8in} M{0.8in}}
\toprule
\multirow{2}{*}{\textbf{Entity}}   & \multicolumn{2}{c}{\textbf{F1}}  \\  \cmidrule{2-3} 
                                & \textbf{any overlap} & \textbf{partial match} \\ \midrule
negation                        & 0.96 & 0.92 \\
region                          & 0.85 & 0.49 \\
side                            & 0.89 & 0.84 \\
size                            & --$^*$   & --$^*$  \\
\bottomrule
\end{tabular}

%% file: tables/agreement_relation.tex
\newcolumntype{M}[1]{>{\centering\arraybackslash}m{#1}}

\begin{tabular}{m{1.0in} M{0.35in}}
\toprule
\textbf{Relation pair} & \textbf{F1}  \\ \midrule
region-negation         & 0.96 \\
region-side             & 0.76 \\
region-size             & --$^*$   \\
\bottomrule
\end{tabular}

%% file: sections/results.tex
HANSO is trained on the PAC training partition and evaluated here using two approaches: \textit{text-versus-text} and \textit{text-versus-image}. In the \textit{text-versus-text} approach, the trained HANSO model is evaluated on the withheld PAC test set. The \textit{text-versus-text} approach is a typical NLP performance evaluation, where the model is trained and evaluated on annotated text. In the \textit{text-versus-image} approach, we apply HANSO to a set of chest radiograph reports for which the associated chest radiograph images are directly annotated by an expert radiologist. The \textit{text-versus-image} approach compares labels derived from text reports against gold standard image annotations. 

\subsection*{Text-versus-text}









\begin{table}[ht]
    \small
    \centering
    \begin{minipage}[c]{3.5in}
        \begin{subtable}[h]{3.2in}

\input{tables/performance_infiltrates}

        \caption{Infiltrates}
        \label{performance_infiltrates}
    \end{subtable}
    
    \vspace{0.2in}
    
    \begin{subtable}[h]{3.2in}

\input{tables/performance_extraparenchymal}

        \caption{Extraparenchymal}
        \label{performance_extraparenchymal}        
    \end{subtable}
    
    \vspace{0.1in}
    
    \caption{Document label prediction performance and gold standard label counts. $^\dag$ indicates the best performing model with significance ($p < 0.05$).}
        \label{performance_doc}    
    \end{minipage}
\hfill
    \begin{minipage}[c]{2.6in}
        \begin{center}
        \includegraphics[width=2.5in]{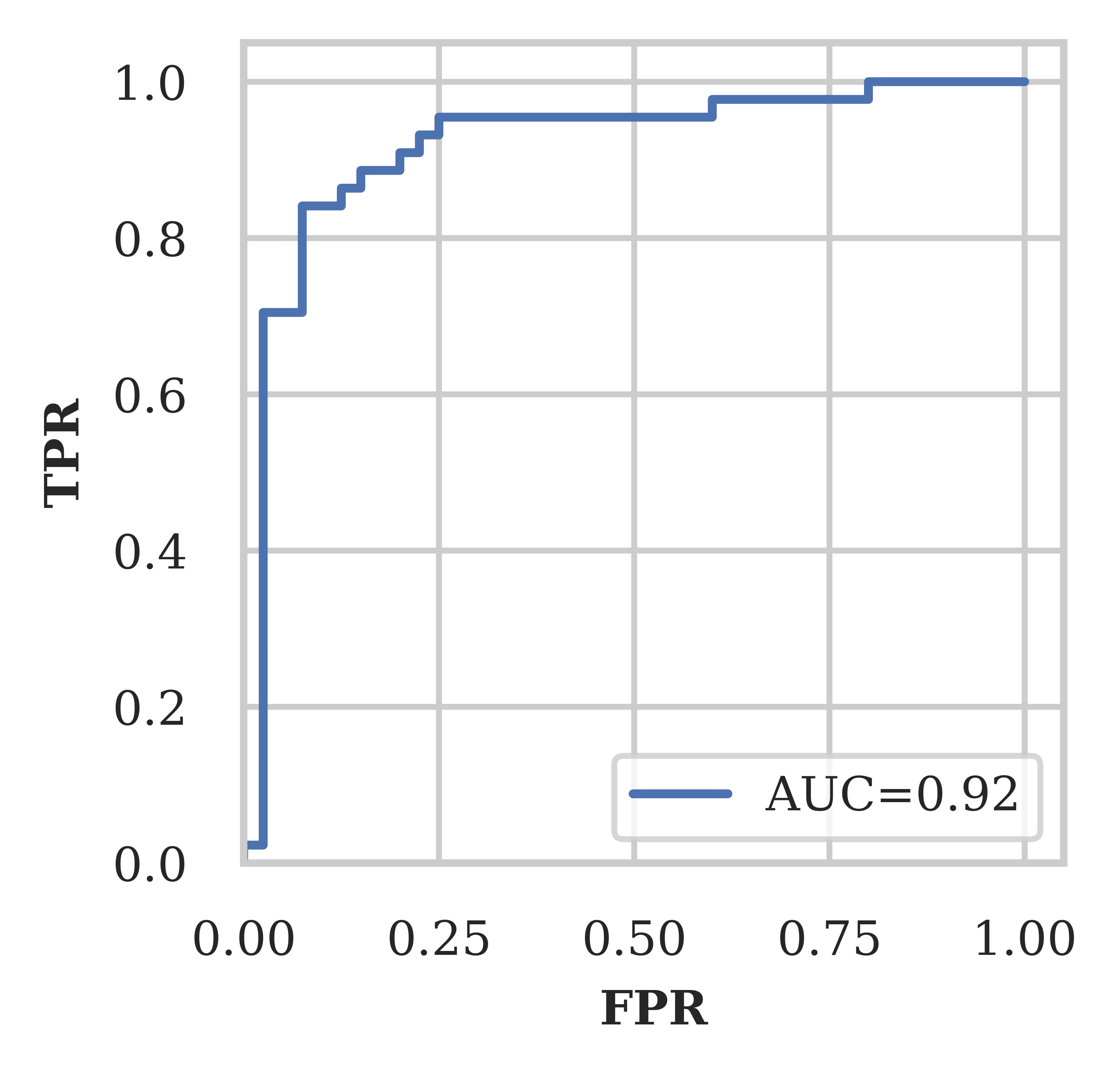}
        \end{center}
        \captionof{figure}{ROC for binary prediction of \textit{infiltrates} as \textit{bilateral} and \textit{not bilateral}}
        \label{roc}
  \end{minipage}
\end{table}

In this subsection, we evaluate model performance on the withheld PAC test set. We present performance results for three models: SVM, \textit{HANSO lite}, and \textit{HANSO full}. To account for the variance associated with model random initialization, each model is trained on the training set 10 times and evaluated on the test set to generate a distribution of performance values. Table \ref{performance_doc} presents the average performance across the 10 runs for the document labels, \textit{infiltrates} and \textit{extraparenchymal}. Performance is presented for each label and the micro average across labels (``micro''). \textit{HANSO full} achieves the best overall performance (F1 micro) for both \textit{infiltrates} and \textit{extraparenchymal} with significance ($p < 0.05$), demonstrating that the inclusion of the sentence objectives contributes to document prediction performance. \textit{HANSO full} achieves a statistically significant improvement in identifying \textit{bilateral} \textit{infiltrates} relative to the SVM ($p<0.01$), and the improvement of \textit{HANSO full} over \textit{HANSO lite} barely misses significance criteria ($p=0.07$). Significance is assessed using a two-side t-test with unequal variance. The average performance across all sentence-level tasks in the \textit{HANSO full} runs is 0.83 F1. Considering the sentence-level tasks represent the summarization of the most salient relation information without including any span information, this performance is high.


To assess the receiver operating characteristic (ROC) for HANSO's ability to identify bilateral infiltrates, we train a separate HANSO model (HANSO full) with binary document targets ($0=not\mbox{ }bilateral$ and $1=bilateral$), where the \textit{none}, \textit{present}, and \textit{unilateral} labels are mapped to \textit{not bilateral}. This binary variant achieves similar performance (0.88 F1) in identifying bilateral infiltrates as the multi-class models in Table \ref{performance_doc}. Figure \ref{roc} presents the ROC for bilateral infiltrates identification using the binary HANSO model. The ROC area under the curve (AUC) is 0.92. Optimizing the prediction threshold using Youden's J statistic yields $J=0.77$ at $FPR=0.08$ and $TPR=0.84$.


\subsection*{Text-versus-image}
\begin{wraptable}{R}{2.7in}
    \small
    \centering

    \begin{subtable}[h]{2.6in}    
    \centering

\input{tables/image_manual}
    \caption{Manual report labels vs. image labels}
    \label{manual_anno}
    \end{subtable}

    \vspace{0.2in}

    \begin{subtable}[h]{2.6in}    
    \centering

\input{tables/image_auto}
    \caption{HANSO report labels vs. image labels}
    \label{auto_anno}        
    \end{subtable}

    \vspace{0.1in}

    \caption{Comparison of radiograph image labels and radiograph report labels for \textit{infiltrates}}
    \label{image_anno}        

\end{wraptable}

In this subsection, we compare the manual PAC annotations and automatically generated HANSO labels against annotated chest radiograph images. An expert radiologist annotated 154 chest radiograph images from \textit{Data set A} with quadrant-level consolidation scores that can be mapped to the \textit{infiltrates} labels \textit{none}, \textit{unilateral}, and \textit{bilateral}, which we treat as the gold standard labels in this section. The annotated radiograph images correspond with 44 annotated reports in PAC (35 train and 9 test). The manual report labels are evaluated for the 44 reports in PAC that have a corresponding annotated images. The HANSO labels are evaluated for the 119 (154-35) annotated radiograph images not associated with reports in the PAC train set. Table \ref{image_anno} presents the performance of the manual and HANSO \textit{infiltrates} labels. While the performance for \textit{none} and \textit{unilateral} is lower for both the manual and HANSO labels, the performance for \textit{bilateral} is high for both the manual labels (0.84 F1) and HANSO labels (0.87 F1).  ARDS diagnosis requires the presence of bilateral infiltrates, so the \textit{bilateral} label performance is most important. HANSO achieves a sensitivity (recall) of 0.85 and specificity of 0.75 for \textit{bilateral} in a one-versus-rest evaluation (\textit{bilateral} vs. \textit{not \mbox{} bilateral}). As the manual and HANSO performance is assessed using different samples, the significance of the performance differences cannot be assessed.

\ifsubfile
\bibliography{mybib}
\fi

%% file: tables/performance_infiltrates.tex


\newcolumntype{M}[1]{>{\centering\arraybackslash}m{#1}}

\begin{tabular}{m{0.63in} M{0.30in} M{0.30in} M{0.35in} M{0.35in} M{0.30in}}
\toprule
\multirow{2}{*}{\textbf{Model}}    & \multicolumn{5}{c}{\textbf{F1}}                               \\ \cmidrule{2-6}
                              & none  & present         & unilateral        & bilateral       & micro \\ \midrule
SVM                           & 0.86  & 0.00            & 0.35              & 0.81            & 0.75        \\
HANSO lite                       & 0.73  & 0.12            & 0.47              & 0.82            & 0.71        \\
HANSO full                       & 0.84  & 0.15            & 0.69$^{\dag}$    & 0.87            & 0.79$^{\dag}$        \\ \midrule  
\# Gold                     & 20                & 6                 & 14                    & 44                    & 84        \\ \bottomrule
\end{tabular}

%% file: tables/performance_extraparenchymal.tex
\newcolumntype{M}[1]{>{\centering\arraybackslash}m{#1}}

\begin{tabular}{m{0.63in} M{0.30in} M{0.30in} M{0.35in} M{0.35in} M{0.30in}}
\toprule
\multirow{2}{*}{\textbf{Model}}         & \multicolumn{5}{c}{\textbf{F1}}                               \\ \cmidrule{2-6}
                              & none & present  & unilateral    & bilateral     & micro \\ \midrule
SVM                           & 0.80 & 0.00     & 0.48          & 0.27          & 0.68              \\
HANSO lite     & 0.86 & 0.03     & 0.57          & 0.61          & 0.74              \\
HANSO full                      & 0.90 & 0.03     & 0.71$^\dag$  & 0.71$^\dag$    & 0.80$^\dag$  \\ \midrule
\# Gold                     & 49                & 2                 & 15                    & 18                    & 84        \\ \bottomrule

\end{tabular}

%% file: tables/image_manual.tex
\begin{tabular}{l c c c c}
\toprule
\textbf{Label}      & \textbf{\# Gold}    & \textbf{P}     & \textbf{R}     & \textbf{F1}   \\ \midrule
none       &    9       & 0.83  & 0.56  & 0.67 \\
unilateral &    3       & 0.17  & 0.33  & 0.22 \\
bilateral  &    32      & 0.87  & 0.81  & 0.84 \\ \bottomrule
\end{tabular}

%% file: tables/image_auto.tex
\begin{tabular}{l c c c c}
\toprule
\textbf{Label}      & \textbf{\# Gold}    & \textbf{P}     & \textbf{R}     & \textbf{F1}    \\ \midrule
none       & 23         & 0.58  & 0.48  & 0.52   \\
unilateral & 13         & 0.33  & 0.38  & 0.36   \\
bilateral  & 83         & 0.89  & 0.86  & 0.87   \\ \bottomrule
\end{tabular}

%% file: sections/conclusions.tex
We introduce a new annotated corpus of chest radiograph reports, PAC, which includes document, entity, and relation annotations associated with ARDS. We also introduce the multi-task, end-to-end HANSO classification framework, which hierarchically encodes documents by encoding the word in sentences and the sentences in a documents. Inter-annotator agreement for the PAC document labels is high (0.90 F1). The agreement for the entities indicates the annotators are generally identifying the same phenomenon in the chest radiograph reports (``any overlap'' agreement 0.85-0.96 F1), although there is variability in the bounds of the annotated spans for the entities (``partial match'' agreement 0.49-0.92 F1). The annotation scheme defines entities with type and subtype labels that normalize each span and capture the information most relevant to ARDS, so the variability in the bounds of the span annotations does not materially impact the clinical meaning of the annotations. However, this variability makes span extraction (entity recognition) challenging. To leverage the entity and relation annotations, we introduce an approach for mapping relations to a one-hot encoding of the entity pairs in each relation and use this one-hot encoding of the relations to create a set of sentence classification tasks. The one-hot encoding captures the most important annotated relation information, without requiring the prediction of entity spans. The primary objective of the HANSO framework is the prediction of the document labels associated with ARDS; however, HANSO includes a secondary objective associated with the prediction of the sentence-level one-hot encodings of the relations. The inclusion of the sentence-level objective increases performance, with significance, for the document labels; \textit{infiltrates} increases from 0.71 to 0.79 F1, and \textit{extraparenchymal} increases from 0.74 to 0.80 F1. The presence of bilateral infiltrates is predicted with very high performance (0.87 F1). HANSO predicts the one-hot encoded relations with high performance (0.83 F1), indicating the model is able to identify the key information from the fine grained annotations without explicit knowledge of span information. HANSO also outperforms a strong SVM baseline from recent ARDS information extraction work, with significance. We also assess the performance of manual (human) and automatic HANSO chest radiograph reports labels relative to annotated chest radiograph images. In the identification of bilateral infiltrates, HANSO achieves high performance against the annotated images (0.87 F1), which is comparable to the human performance (0.84 F1).

ARDS is a common complication of COVID-19 infection with high mortality; however, the identification of ARDS is often delayed or missed entirely. Delays in diagnosis lead to delays in evidence-based therapies that can improve clinical outcomes in COVID-associated ARDS. To our knowledge, this is the first study in COVID-19 that uses radiology reports to develop an automated NLP algorithm for identifying ARDS. These algorithms could be implemented in EHRs for the real-time surveillance of ARDS in COVID-19 infected populations with the goal of ensuring early implementation of evidence-based strategies for decreasing ARDS mortality. As next steps, we will complete an external validation of the data set and modeling framework to prepare for the deployment of an ARDS diagnostic tool. Specifically, the developed HANSO algorithm and trained model will be incorporated into a larger diagnostic tool that will be released and validated within the Electronic Medical Records and Genomics (eMERGE) Network to support pulmonary phenotyping efforts across different sites.

\ifsubfile
\bibliography{mybib}
\fi